\documentclass[runningheads]{llncs}
\usepackage{eccv}
\usepackage{eccvabbrv}
\usepackage{graphicx}
\usepackage{booktabs}
\usepackage{multirow}
\usepackage[table]{xcolor}
\usepackage{microtype}
\usepackage[accsupp]{axessibility}  
\usepackage{hyperref}
\usepackage{orcidlink}

\begin{document}
\title{BELDE: Building a Large-scale Earth-observation Land-cover Dataset for Europe}
\titlerunning{BELDE}

\author{Ümit Mert Çağlar\orcidlink{0000-0002-0391-3847} \and
Alptekin Temizel\orcidlink{0000-0001-6082-2573}}

\authorrunning{Ü.M.~Çağlar And A.~Temizel}

\institute{Graduate School of Informatics, Middle East Technical University, Ankara, Turkey
\email{\{mert.caglar,atemizel\}@metu.edu.tr}}
\maketitle

\begin{abstract}
Earth observation imagery plays a critical role in environmental monitoring, urban planning, disaster assessment, and climate analysis. While multi-spectral sensors are increasingly available, true-color (RGB) imagery remains widely used due to the power, cost, and deployment constraints of many satellite and aerial platforms. However, existing land-cover segmentation datasets are often limited in geographic coverage, scale, or public accessibility.

To bridge this gap, we introduce BELDE (Building a Large-scale Earth-observation Land-cover Dataset for Europe), a publicly available dataset tailored for RGB-based remote sensing semantic segmentation. Constructed from Sentinel-2 true-color images and ESA WorldCover data annotations, BELDE contains 1,088,385 curated image-segmentation map pairs spanning Europe with 7 land-cover classes at 10 m spatial resolution, making it one of the largest publicly available RGB land-cover segmentation datasets for Earth observation. To facilitate cross-region generalization studies, we additionally introduce BELDE-K (16,607 pairs) covering the Republic of Korea and BELDE-CA-NV (88,155 pairs) covering California and Nevada in the United States.

We establish baseline results using multiple semantic segmentation architectures and evaluate both in-domain and cross-domain performance. Models trained on BELDE achieve an F1 score of 83.0\% on the European test set, while performance decreases to 66.4\% on BELDE-CA-NV and 58.3\% on BELDE-K, highlighting the challenges posed by out-of-distribution geographic domain shift. By providing a continental-scale RGB segmentation and evaluation benchmark, BELDE supports the development of robust and transferable Earth observation models. The dataset and benchmark resources will be publicly released.
\keywords{Remote sensing \and Earth observation \and Segmentation \and Dataset}
\end{abstract}

\section{Introduction}
\label{sec:introduction}
Accurate land-cover semantic segmentation is a fundamental capability for Earth Observation (EO), enabling applications such as environmental monitoring, climate assessment, urban planning, precision agriculture, and disaster response~\cite{audebert2016semantic}. The increasing availability of satellite imagery, particularly from programs such as ESA’s Sentinel-2 mission, has driven significant advances in deep learning approaches for large-scale EO analysis~\cite{ma2019deep,demir2018deepglobe}. However, the success of modern segmentation models remains highly dependent on the availability of large, diverse, and publicly accessible annotated datasets.

Although multi-spectral imagery can improve segmentation performance, RGB imagery remains one of the most widely available sensing modalities across satellite, aerial, and archival Earth observation platforms. RGB imagery is readily accessible, requires lower storage and transmission bandwidth, and is compatible with a broad range of operational and resource-constrained sensing systems. Consequently, RGB-based segmentation models remain highly relevant for real-world deployment and large-scale Earth observation applications.

Despite substantial progress in remote sensing datasets, existing semantic segmentation benchmarks often suffer from one or more limitations: restricted geographic coverage, limited dataset scale, narrow environmental diversity, or constrained public accessibility~\cite{sumbul2019bigearthnet}. As a result, models trained on these datasets frequently exhibit degraded performance when used for regions with different climates, land-cover distributions, atmospheric conditions, or acquisition characteristics~\cite{al2025analysing}. This lack of geographically diverse large-scale benchmarks hinders the development and evaluation of robust Earth observation models capable of generalizing beyond their training domains.

\begin{figure}[htbp]
    \centering
    \includegraphics[trim={0 12cm 0 0}, clip, width=\linewidth]{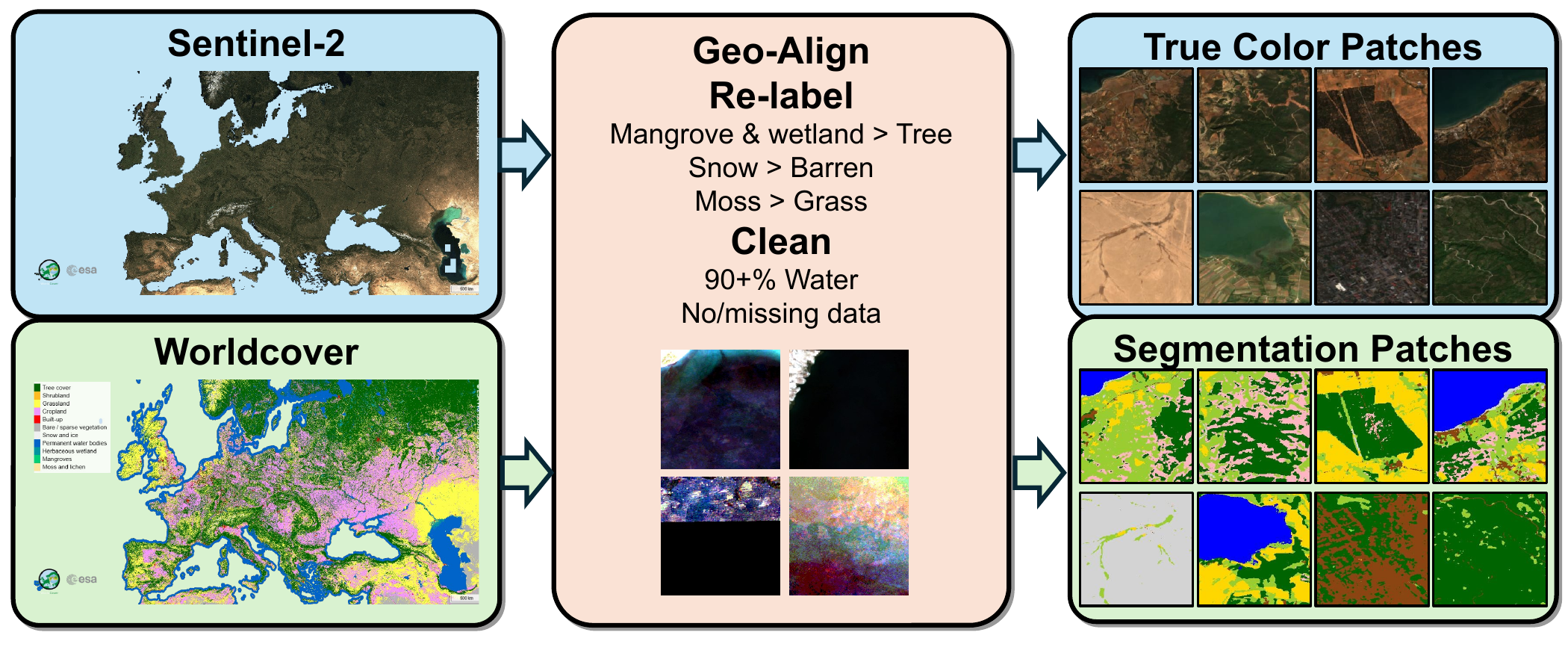}
    \caption{Workflow of the BELDE dataset generation. Automated spatial querying downloads co-registered Sentinel-2 true-color tiles and ESA WorldCover 2021 maps. The pipeline applies rule-based taxonomic remapping, strict no-data filtering, and spatial patch slicing to construct over one million curated $256 \times 256$ pixel image-mask pairs.}
    \label{fig:pipeline_overview}
\end{figure}

To address these limitations, we introduce \textbf{BELDE} (\textbf{B}uilding a Large-scale \textbf{E}arth-observation \textbf{L}and-cover \textbf{D}ataset for \textbf{E}urope), a publicly available benchmark for RGB remote sensing semantic segmentation. Constructed from Sentinel-2 True Color Imagery (TCI) and ESA WorldCover annotations (Figure~\ref{fig:pipeline_overview}), BELDE contains more than 1.08 million curated image-mask pairs spanning a geographically diverse region covering much of Europe. The dataset encompasses a broad range of climates, ecosystems, agricultural regions, urban environments, forests, and coastal areas within the geographic extent of $13^\circ \text{W}$ to $50^\circ \text{E}$ longitude and $33^\circ \text{N}$ to $60^\circ \text{N}$ latitude. By focusing exclusively on RGB imagery while maintaining continental-scale coverage, BELDE provides a large-scale benchmark for representation learning and semantic segmentation in realistic Earth observation settings. We use non-overlapping patches throughout this work to prevent data-leakage and overfitting as discussed in recent studies, highlighting serious issues with public datasets containing duplicate entries in both train and test sets~\cite{adimoolam2026data}.

Beyond dataset scale, an important challenge in Earth observation is understanding how well models generalize across geographic regions. To facilitate systematic evaluation of domain shift, we additionally introduce two geographically distinct benchmark extensions: \textbf{BELDE-K}, covering the Republic of Korea, and \textbf{BELDE-CA-NV}, covering California and Nevada in the United States. These datasets enable controlled out-of-distribution evaluation and provide a benchmark for studying the transferability of segmentation models trained on European imagery.

BELDE is generated through a fully automated and reproducible processing pipeline that combines Sentinel-2 imagery with ESA WorldCover land-cover maps. The pipeline performs spatial alignment, taxonomic harmonization, quality filtering, and patch extraction to produce training-ready image-mask pairs at 10-meter spatial resolution. The complete preprocessing framework is released alongside the dataset to promote transparency, reproducibility, and future dataset expansion.

The contributions of this work are summarized as follows:

\begin{itemize}
\item We introduce BELDE, a large-scale publicly available RGB remote sensing semantic segmentation dataset containing more than 1.08 million geo-aligned image-mask pairs across Europe.
\item We also release two complementary out-of-distribution benchmarks, BELDE-K and BELDE-CA-NV, enabling systematic evaluation of geographic domain shift and model transferability.
\item We establish baseline results using multiple state-of-the-art semantic segmentation architectures and quantify the challenges of cross-region generalization in large-scale land-cover mapping.
\item We provide a fully automated, reproducible, and open-source dataset generation pipeline, facilitating future dataset maintenance, verification, and extension.
\end{itemize}

\section{Related Work}
\label{sec:related_work}
Large-scale land-cover products such as ESA WorldCover~\cite{zanaga2022esa} and Dynamic World~\cite{brown2022dynamic} provide globally consistent land-cover annotations derived from Sentinel-2 imagery. These resources have become important foundations for Earth observation research; however, they are primarily distributed as geospatial products rather than machine-learning-ready semantic segmentation benchmarks.

Several curated remote sensing datasets have been introduced to facilitate deep learning research. EuroSAT~\cite{helber2019eurosat} and BigEarthNet~\cite{clasen2025reben} provide large collections of Sentinel-2 imagery but focus on scene-level classification and multi-label categorization. For semantic segmentation, datasets such as DeepGlobe~\cite{demir2018deepglobe}, LandCover.ai~\cite{Boguszewski_2021_CVPR}, and LoveDA~\cite{BENCHMARKS2021_4e732ced} have provided valuable benchmarks; however, they remain geographically restricted and comparatively limited in scale.

More recently, YieldSAT~\cite{Miranda_2026_CVPR} introduces a high-resolution, multimodal benchmark dataset for pixel-level crop yield using multi-spectral satellite imagery (Sentinel-2) across multiple crops and climate zones. Similarly JAXA high-resolution land-use and land-cover map provides segmentation maps for Japan~\cite{hirayama2022generation} and Vietnam~\cite{truong2024jaxa}. Recent study also delves into spatial dependencies between neighboring Earth observation images~\cite{Zeng_2026_CVPR} indicating patches in close proximity can be exploited to improve model training. Conversely, data leakage and duplicate imagery in training and testing sets have been identified in a popular Earth observation dataset, creating overfitting problem~\cite{adimoolam2026data}. On the multi-modal part, GeoViS enables visual search for remote sensing with visual grounding~\cite{Zhang_2026_CVPR}, utilizing visual retrieval augmented generation (RAG). In the field of synthetic data augmentation, ARAS400k~\cite{ccauglar2026grounding} investigated dataset scaling expanding a corpus of real Sentinel-2 image and land-cover mask pairs with generated samples and accompanying vision-language annotations. While this demonstrated the effectiveness of synthetic data augmentation, the dataset remains focused on a single geographic region. In contrast, BELDE emphasizes large-scale curation of real RGB imagery across a geographically diverse European footprint, resulting in over one million image-mask pairs and enabling systematic studies of domain shift. In comparison with other publicly available datasets, available in Table~\ref{tab:dataset_comparison}, BELDE provides one of the largest curated RGB Earth observation imagery with associated land-cover segmentation maps.

\begin{table}[htbp]
\centering
\caption{Comparison of remote sensing datasets.}
\label{tab:dataset_comparison}
\begin{tabular}{lrrccc}
\toprule
\textbf{Dataset} & \textbf{Patches} & \textbf{Classes} & \textbf{Resolution} & \textbf{Patch Size} & \textbf{Multi-spectral} \\
\midrule
BigEarthNet & 549,488 & 19 & 10m & $120 \times 120$ & No \\
ARAS (Real) & 100,240 & 7 & 10m & $256 \times 256$ & No \\
EuroSAT & 27,000 & 10 & 10m & $64 \times 64$ & Yes \\
LoveDA & 5,987 & 7 & 0.3m & $1024 \times 1024$ & No \\
DeepGlobe & 1,146 & 7 & 0.5m & $2448 \times 2448$ & Yes \\
LandCover.ai & 10,674 & 4 & 0.5m & $512 \times 512$ & No \\
YieldSAT & 113,555 & 12 & 20m & Varied & Yes \\
\midrule
BELDE (ours) & 1,088,385 & 7 & 10m & $256 \times 256$ & No \\
BELDE-K (ours) & 16,607 & 7 & 10m & $256 \times 256$ & No \\
BELDE-CA-NV (ours) & 88,155 & 7 & 10m & $256 \times 256$ & No \\
\bottomrule
\end{tabular}
\end{table}

\subsection{Semantic Segmentation Architectures for Earth Observation}

Modern Earth observation segmentation systems are predominantly based on encoder-decoder CNNs, vision transformers, or hybrid architectures. Representative CNN-based approaches include UNet~\cite{ronneberger2015u}, UNet++\cite{zhou2018unet++}, LinkNet\cite{chaurasia2017linknet}, and DeepLabV3+\cite{chen2018encoder}. Transformer-based models such as SegFormer\cite{xie2021segformer} and DPT~\cite{ranftl2021vision} have demonstrated strong performance by capturing long-range spatial dependencies, while recent efficiency-oriented architectures including EfficientFormer~\cite{li2023rethinking}, FastViT~\cite{vasu2023fastvit}, and LALE~\cite{ccauglar2026lale} aim to balance accuracy and computational cost for deployment on resource-constrained platforms.

Rather than proposing a new segmentation architecture, BELDE provides a large-scale benchmark for evaluating these diverse model families under both in-domain and cross-domain settings. Accordingly, our experiments include representative CNN, transformer, and lightweight hybrid architectures to characterize the performance-efficiency trade-offs of modern segmentation models on large-scale RGB Earth observation imagery.
\section{Dataset Acquisition and Curation Pipeline}
\label{sec:dataset_pipeline}

\begin{figure}[htbp]
    \centering
    \includegraphics[trim={0 0 0 0}, clip, width=1\linewidth]{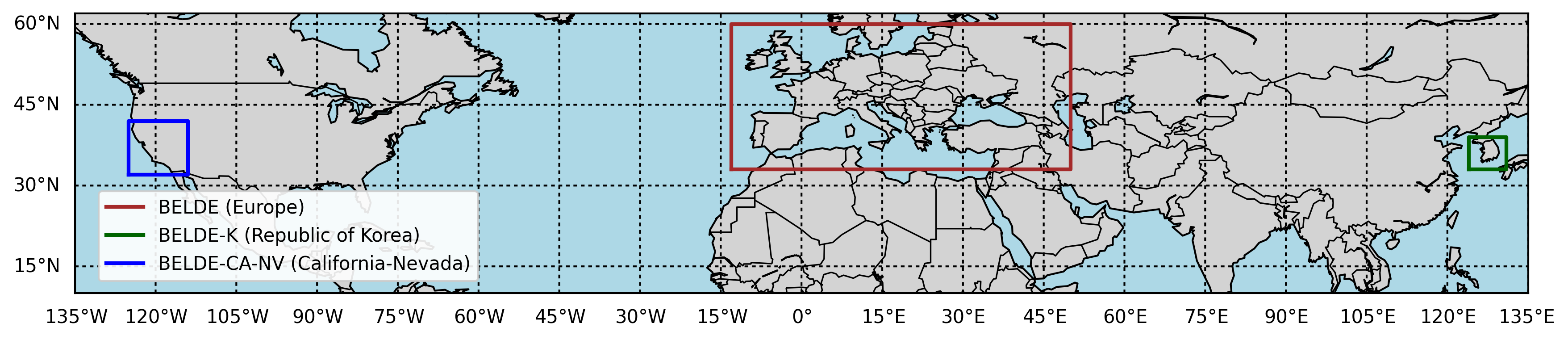}
    \caption{The data acquisition area of the BELDE, BELDE-K and BELDE-CA-NV corresponding to Europe, Republic of Korea and California-Nevada respectively.}
    \label{fig:eu_bbox}
\end{figure}

\begin{figure}[htbp]
    \centering
    \includegraphics[width=0.75\linewidth]{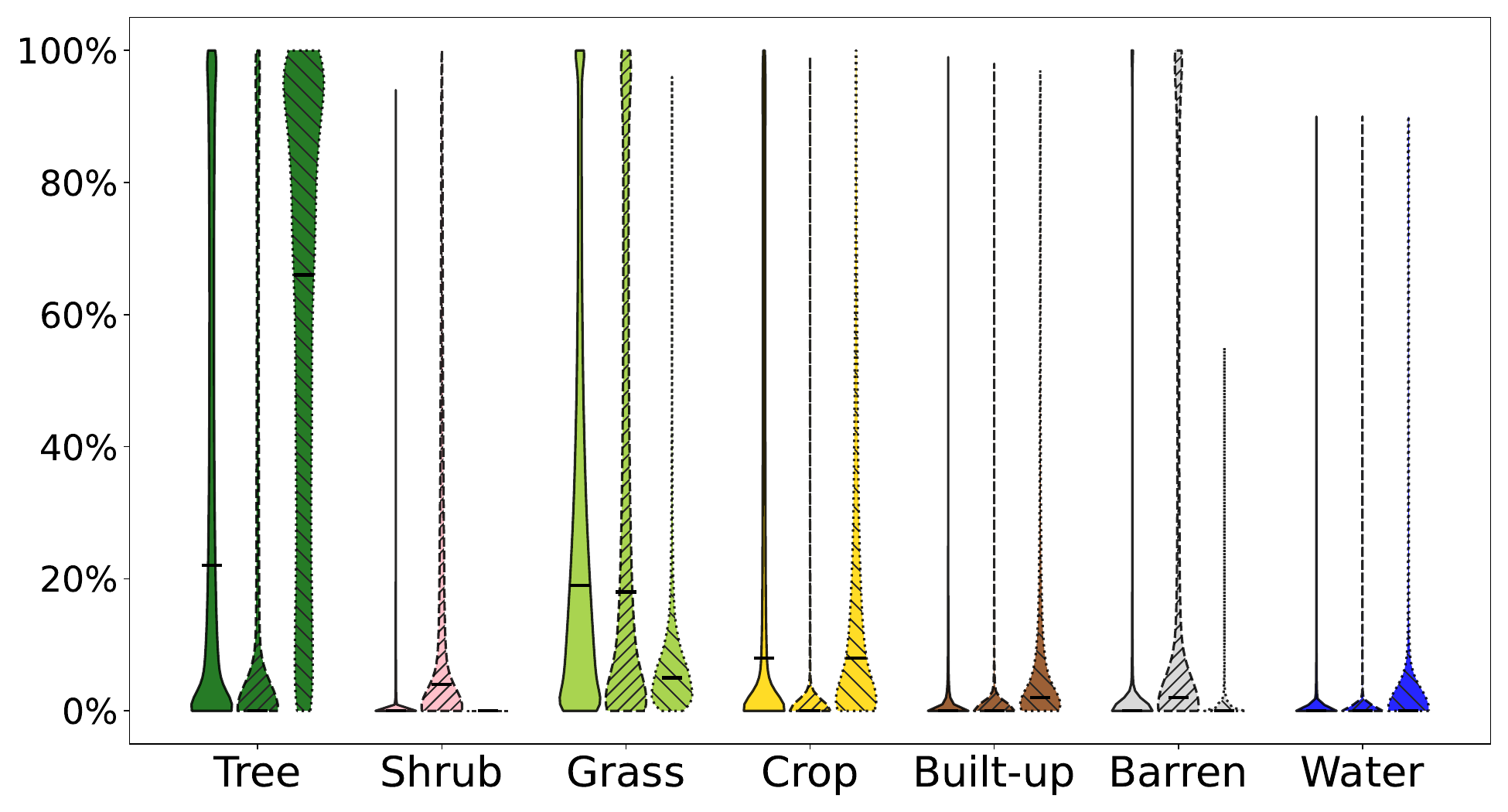}
    \caption{Class composition violin plots for BELDE (solid), BELDE-CA-NV (forward slash hatched) and BELDE-K (backward slash hatched) in order}
    \label{fig:composition}
\end{figure}

BELDE is constructed through a fully automated and reproducible pipeline that generates co-registered image-mask pairs for semantic segmentation from large-scale Earth observation data (Figure~\ref{fig:pipeline_overview}). Given a user-defined geographic region, the pipeline retrieves Sentinel-2 true-color imagery and corresponding ESA WorldCover annotations, and processes them into training-ready RGB segmentation patches.

\subsection{Data Acquisition}

We construct BELDE using Sentinel-2 true-color imagery (TCI) and ESA WorldCover 2021 land-cover products~\cite{zanaga2022esa}. WorldCover product provides the RGB images as annual composites, a set of yearly median and percentiles composites at 10m resolution, enabling efficient cloud removal and normalization of the RGB images. The acquisition region spans Europe, bounded by $13^\circ\text{W}$-$50^\circ\text{E}$ longitude and $33^\circ\text{N}$-$60^\circ\text{N}$ latitude (Figure~\ref{fig:eu_bbox}). This region captures a wide range of ecological and climatic conditions, including Mediterranean, continental, oceanic, boreal and alpine environments. 

For each spatial location, co-registered RGB imagery and land-cover labels are retrieved and aligned through their geospatial identifiers. The pipeline is fully automated and designed to ensure robustness against interrupted downloads, corrupted files and inconsistent tile retrieval across large-scale spatial queries.

\subsection{Raster Alignment and Patch Extraction}

Sentinel-2 imagery and WorldCover annotations are distributed as large georeferenced raster tiles. To construct machine-learning-ready samples, corresponding image and label rasters are spatially aligned and partitioned into fixed-size patches. We apply a sliding-window strategy to generate non-overlapping $256 \times 256$ pixel patches at 10-meter spatial resolution.

To ensure spatial consistency, segmentation masks are resampled using nearest-neighbor interpolation, preserving categorical label integrity during alignment.

\subsection{Label Harmonization}

The original ESA WorldCover product defines 11 land-cover classes~\cite{zanaga2022esa}. However, several classes are rare, spatially localized, or exhibit semantic overlap under RGB observation conditions. To improve class balance and reduce label ambiguity, we apply a rule-based class aggregation strategy.

Specifically, herbaceous wetland and moss/lichen are merged into grassland, mangroves are merged into tree cover, and snow/ice is merged into barren land. After harmonization, BELDE contains 7 semantic classes: tree, shrub, grass, crop, built-up, barren, and water.

This formulation aligns the dataset with commonly used land-cover taxonomies and improves training stability for RGB-based segmentation models.

\subsection{Quality Filtering}

To ensure high-quality training data, we apply two filtering criteria. First, we remove any patch containing missing values in either the RGB image or the segmentation mask. Second, we exclude patches where more than 90\% of pixels correspond to open water, to prevent severe class imbalance and over-representation of homogeneous regions.

In total, the filtering process removes: 480,116 samples due to missing or corrupted RGB data, 1,380,584 samples due to missing or invalid land-cover labels and 81,027 samples due to excessive water coverage. After filtering, the dataset contains 1,088,385 valid image-mask pairs covering $7{,}132{,}839~\mathrm{km}^2$.

\subsection{Dataset Composition and Splits}

The final dataset with non-overlapping patches, is split into training, validation, and test sets using an 80\% / 10\% / 10\% stratified split, preserving overall class distributions across partitions. The resulting class proportions are approximately: 35\% tree, 1\% shrub, 28\% grass, 26\% crop, 2\% built-up, 6\% barren, and 2\% water.

In addition to the European dataset, we construct two geographically distinct extensions for cross-region evaluation: BELDE-CA-NV (California and Nevada, USA) and BELDE-K (Republic of Korea). These subsets are generated using the same pipeline to ensure consistency while enabling controlled evaluation of geographic domain shift.

\subsection{Dataset Storage and Access}

For efficient storage and fast loading during benchmarking, BELDE is provided in a compressed columnar format using Apache Parquet, and accessed via Apache Arrow, enabling low-overhead data retrieval with near zero-copy access.

\section{Semantic Segmentation Results}

We evaluate 17 semantic segmentation model configurations (Table~\ref{tab:overall_metrics}) spanning three architectural families: (i) encoder-decoder networks, (ii) transformer-based dense prediction models and (iii) lightweight efficiency-oriented architectures. This includes Encoder-decoder models with CNN-backbones: DeepLabV3~\cite{chen2017rethinking}, DeepLabV3+~\cite{chen2018encoder}, FPN~\cite{lin2017feature}, LinkNet~\cite{chaurasia2017linknet}, PSPNet~\cite{zhao2017pyramid}, UNet~\cite{ronneberger2015u}, UNet++\cite{zhou2018unet++} and Transformer-backbones SegFormer\cite{xie2021segformer}, DPT variants such as DeiT3~\cite{touvron2022deit} and MaxViT~\cite{tu2022maxvit}, while efficiency-oriented architectures include EfficientFormer~\cite{li2023rethinking}, FastViT~\cite{vasu2023fastvit}, and LALE~\cite{ccauglar2026lale}. Across all configurations, model sizes range from 1.5M to 117M parameters, enabling evaluation across a broad efficiency-accuracy spectrum.

\subsection{Training Protocol}

To ensure fair comparison, all models are trained under an identical optimization pipeline. We use the AdamW optimizer with an initial learning rate of $1 \times 10^{-4}$ and optimize a multiclass Dice loss objective. The learning rate is reduced based on validation F1-score using a plateau-based scheduling strategy. Input images are normalized using ImageNet statistics and augmented with horizontal and vertical flips, $90^\circ$ rotations, and color jittering. Each model is trained for up to 10 epochs with early stopping based on validation F1-score.

\begin{table}[ht]
\centering
\caption{Overall evaluation metrics (\%) grouped by architectural cohorts.}
\label{tab:overall_metrics}
\renewcommand{\arraystretch}{1.1}
\resizebox{0.9\textwidth}{!}{\small\begin{tabular}{l|c|c|c|c|c|c}
Architecture & Accuracy & Precision & Recall & F1-score & IoU & Parameters\\
\midrule
\rowcolor{gray!12}
DeepLabV3 & $86.4 \pm 0.1$ & $79.2 \pm 0.3$ & $81.6 \pm 0.2$ & $80.2 \pm 0.1$ & $69.2 \pm 0.1$ & 33.06M\\
\rowcolor{gray!12}
DeepLabV3+ & $87.0 \pm 0.0$ & $80.7 \pm 0.1$ & $81.8 \pm 0.2$ & $81.2 \pm 0.1$ & $70.4 \pm 0.1$ & 29.49M\\
\rowcolor{gray!12}
FPN & $87.0 \pm 0.1$ & $80.7 \pm 0.2$ & $81.6 \pm 0.3$ & $81.1 \pm 0.2$ & $70.4 \pm 0.2$ & 30.17M\\
\rowcolor{gray!12}
Linknet & $86.7 \pm 0.1$ & $80.6 \pm 0.2$ & $81.6 \pm 0.2$ & $81.0 \pm 0.1$ & $70.2 \pm 0.2$ & 28.75M\\
\rowcolor{gray!12}
PSPNet & $84.3 \pm 0.4$ & $77.4 \pm 0.9$ & $78.8 \pm 0.6$ & $77.9 \pm 0.4$ & $66.2 \pm 0.5$ & 28.44M\\
\rowcolor{gray!12}
Segformer & $87.1 \pm 0.7$ & $80.5 \pm 1.2$ & $81.9 \pm 0.7$ & $81.1 \pm 1.0$ & $70.5 \pm 1.3$ & 28.89M\\
\rowcolor{gray!12}
Unet & $87.5 \pm 0.4$ & $81.3 \pm 0.1$ & $82.5 \pm 0.6$ & $81.8 \pm 0.3$ & $71.3 \pm 0.5$ & 31.22M\\
\rowcolor{gray!12}
Unet++ & $87.4 \pm 0.2$ & $81.1 \pm 0.6$ & $82.5 \pm 0.1$ & $81.7 \pm 0.3$ & $71.2 \pm 0.4$ & 31.91M\\
\midrule
DeiT3 & $87.3 \pm 0.6$ & $81.4 \pm 0.7$ & $81.9 \pm 0.7$ & $81.5 \pm 0.8$ & $71.0 \pm 1.0$ & 117.08M\\
EfficientFormer-L1 & $87.4 \pm 0.5$ & $81.7 \pm 0.8$ & $81.6 \pm 0.6$ & $81.6 \pm 0.6$ & $71.1 \pm 0.8$ & 29.83M\\
EfficientFormer-L3 & $88.0 \pm 0.5$ & $82.4 \pm 0.4$ & $82.8 \pm 1.2$ & $82.5 \pm 0.8$ & $72.3 \pm 0.9$ & 49.13M\\
EfficientFormer-L7 & $88.3 \pm 0.4$ & $82.5 \pm 0.5$ & $83.6 \pm 0.6$ & $83.0 \pm 0.5$ & $72.8 \pm 0.7$ & 100.32M\\
FastViT-mci0 & $87.4 \pm 0.5$ & $81.5 \pm 0.5$ & $82.0 \pm 1.3$ & $81.6 \pm 0.9$ & $71.1 \pm 1.1$ & 29.07M\\
FastViT-sa12 & $87.7 \pm 0.5$ & $81.5 \pm 1.1$ & $83.2 \pm 0.2$ & $82.2 \pm 0.7$ & $71.8 \pm 0.9$ & 29.24M\\
MaxViT & $88.4 \pm 0.4$ & $82.4 \pm 0.8$ & $83.6 \pm 0.2$ & $82.9 \pm 0.6$ & $72.8 \pm 0.7$ & 60.80M\\
\midrule
\rowcolor{gray!12}
LALE-S3 & $84.4 \pm 0.1$ & $78.3 \pm 0.0$ & $78.8 \pm 0.1$ & $78.4 \pm 0.1$ & $66.8 \pm 0.1$ & 3.66M\\
\rowcolor{gray!12}
LALE-S2 & $84.1 \pm 0.3$ & $78.1 \pm 0.2$ & $78.7 \pm 0.2$ & $78.2 \pm 0.2$ & $66.5 \pm 0.2$ & 2.61M\\
\end{tabular}}
\end{table}

\subsection{Overall Performance}

Table~\ref{tab:overall_metrics} reports overall performance across architectures. Transformer-based and hybrid architectures achieve the highest accuracy, with EfficientFormer-L7 and MaxViT reaching F1-scores of 83.0\% and 82.9\%, respectively, albeit with relatively large model sizes (100M and 60M parameters). CNN-based architectures remain competitive, with UNet achieving 81.8\% F1-score at 31M parameters. In contrast, lightweight LALE models achieve a favorable efficiency-performance trade-off, with LALE-S2 reaching 78.2\% F1-score using only 2.6M parameters.

Across all model families, we observe a consistent trade-off between model capacity and segmentation accuracy. Standard deviation across training runs is higher for large transformer-based models (such as DPT), while lightweight architectures (such as LALE) exhibit more stable behavior.

\begin{table}[ht]
\centering
\caption{Per-class F1-scores (\%) grouped by architectural cohorts.}
\label{tab:per_class_f1}
\renewcommand{\arraystretch}{1.1}
\resizebox{0.95 \textwidth}{!}{\begin{tabular}{l|c|c|c|c|c|c|c}
Architecture & Tree & Shrub & Grass & Crop & Built-up & Barren & Water \\
\midrule
\rowcolor{gray!12}
DeepLabV3 & $91.0 \pm 0.2$ & $48.0 \pm 0.6$ & $80.7 \pm 0.1$ & $88.8 \pm 0.0$ & $72.6 \pm 0.6$ & $87.1 \pm 0.1$ & $93.4 \pm 0.1$ \\
\rowcolor{gray!12}
DeepLabV3+ & $91.9 \pm 0.1$ & $49.0 \pm 0.4$ & $81.3 \pm 0.1$ & $89.0 \pm 0.0$ & $76.3 \pm 0.8$ & $87.1 \pm 0.5$ & $93.5 \pm 0.1$ \\
\rowcolor{gray!12}
FPN & $91.8 \pm 0.2$ & $48.8 \pm 0.5$ & $81.1 \pm 0.3$ & $88.9 \pm 0.0$ & $76.2 \pm 1.0$ & $87.4 \pm 0.2$ & $93.6 \pm 0.1$ \\
\rowcolor{gray!12}
Linknet & $91.7 \pm 0.2$ & $48.3 \pm 0.8$ & $80.8 \pm 0.2$ & $88.7 \pm 0.1$ & $77.0 \pm 1.2$ & $87.2 \pm 0.2$ & $93.5 \pm 0.0$ \\
\rowcolor{gray!12}
PSPNet & $89.9 \pm 0.1$ & $42.1 \pm 1.6$ & $77.7 \pm 0.7$ & $86.7 \pm 0.2$ & $71.1 \pm 0.1$ & $85.8 \pm 0.4$ & $91.8 \pm 0.2$ \\
\rowcolor{gray!12}
Segformer & $92.0 \pm 0.6$ & $47.8 \pm 3.1$ & $81.3 \pm 0.9$ & $89.1 \pm 0.6$ & $77.1 \pm 1.4$ & $87.1 \pm 0.8$ & $93.6 \pm 0.4$ \\
\rowcolor{gray!12}
Unet & $92.3 \pm 0.3$ & $49.0 \pm 0.6$ & $81.9 \pm 0.5$ & $89.4 \pm 0.3$ & $78.6 \pm 1.4$ & $87.5 \pm 0.2$ & $93.9 \pm 0.3$ \\
\rowcolor{gray!12}
Unet++ & $92.2 \pm 0.3$ & $49.4 \pm 0.2$ & $81.9 \pm 0.0$ & $89.3 \pm 0.1$ & $77.9 \pm 1.5$ & $87.7 \pm 0.2$ & $93.9 \pm 0.1$ \\
\midrule
DeiT3 & $92.3 \pm 0.4$ & $49.0 \pm 2.0$ & $81.7 \pm 0.8$ & $88.9 \pm 0.5$ & $77.7 \pm 0.9$ & $87.7 \pm 0.4$ & $93.5 \pm 0.4$ \\
EfficientFormer-L1 & $92.3 \pm 0.4$ & $49.3 \pm 1.5$ & $81.6 \pm 0.6$ & $89.1 \pm 0.3$ & $77.7 \pm 0.9$ & $87.6 \pm 0.4$ & $93.7 \pm 0.2$ \\
EfficientFormer-L3 & $92.8 \pm 0.3$ & $51.1 \pm 2.5$ & $82.6 \pm 0.7$ & $89.7 \pm 0.4$ & $79.2 \pm 0.7$ & $88.2 \pm 0.5$ & $94.2 \pm 0.3$ \\
EfficientFormer-L7 & $92.9 \pm 0.3$ & $52.3 \pm 1.5$ & $83.1 \pm 0.5$ & $90.0 \pm 0.3$ & $79.6 \pm 0.6$ & $88.4 \pm 0.4$ & $94.4 \pm 0.2$ \\
FastViT-mci0 & $92.5 \pm 0.4$ & $48.7 \pm 2.4$ & $81.6 \pm 0.9$ & $89.1 \pm 0.5$ & $78.2 \pm 1.0$ & $87.6 \pm 0.5$ & $93.8 \pm 0.5$ \\
FastViT-sa12 & $92.6 \pm 0.3$ & $50.2 \pm 2.1$ & $82.1 \pm 0.9$ & $89.4 \pm 0.4$ & $79.1 \pm 0.5$ & $87.7 \pm 0.4$ & $94.0 \pm 0.4$ \\
MaxViT & $92.9 \pm 0.2$ & $52.0 \pm 1.3$ & $83.2 \pm 0.6$ & $90.2 \pm 0.4$ & $79.5 \pm 0.7$ & $88.5 \pm 0.4$ & $94.4 \pm 0.2$ \\
\midrule
\rowcolor{gray!12}
LALE-S3 & $90.3 \pm 0.1$ & $43.4 \pm 0.2$ & $77.5 \pm 0.2$ & $85.8 \pm 0.2$ & $74.3 \pm 0.0$ & $86.3 \pm 0.2$ & $91.4 \pm 0.1$ \\
\rowcolor{gray!12}
LALE-S2 & $90.2 \pm 0.2$ & $43.3 \pm 0.3$ & $77.4 \pm 0.1$ & $85.6 \pm 0.2$ & $74.2 \pm 0.1$ & $85.9 \pm 0.8$ & $91.1 \pm 0.0$ \\
\end{tabular}}
\end{table}

\subsection{Per-Class Analysis}

Table~\ref{tab:per_class_f1} provides per-class F1-scores across all architectures. We observe consistently strong performance for spectrally homogeneous classes such as tree and water, which exhibit clear visual signatures in RGB imagery. In contrast, more structurally heterogeneous and rare classes, as shown in Figure~\ref{fig:composition}, such as shrub consistently yield lower performance across all models. This difficulty is primarily attributed to the rare occurrence and inter-class visual overlap in RGB observations, meanwhile they are often distinguished by multi-spectral data. As a result, shrub remains the most challenging class across all evaluated architectures.

Transformer-based models show improved performance on structurally complex classes compared to encoder-decoder baselines with CNN backbones, suggesting better global context aggregation. Meanwhile, lightweight LALE models provide competitive performance with significantly reduced parameter counts.

\begin{figure}
    \centering
    \includegraphics[trim={1.4cm 0 0 0}, clip, width=0.8\linewidth]{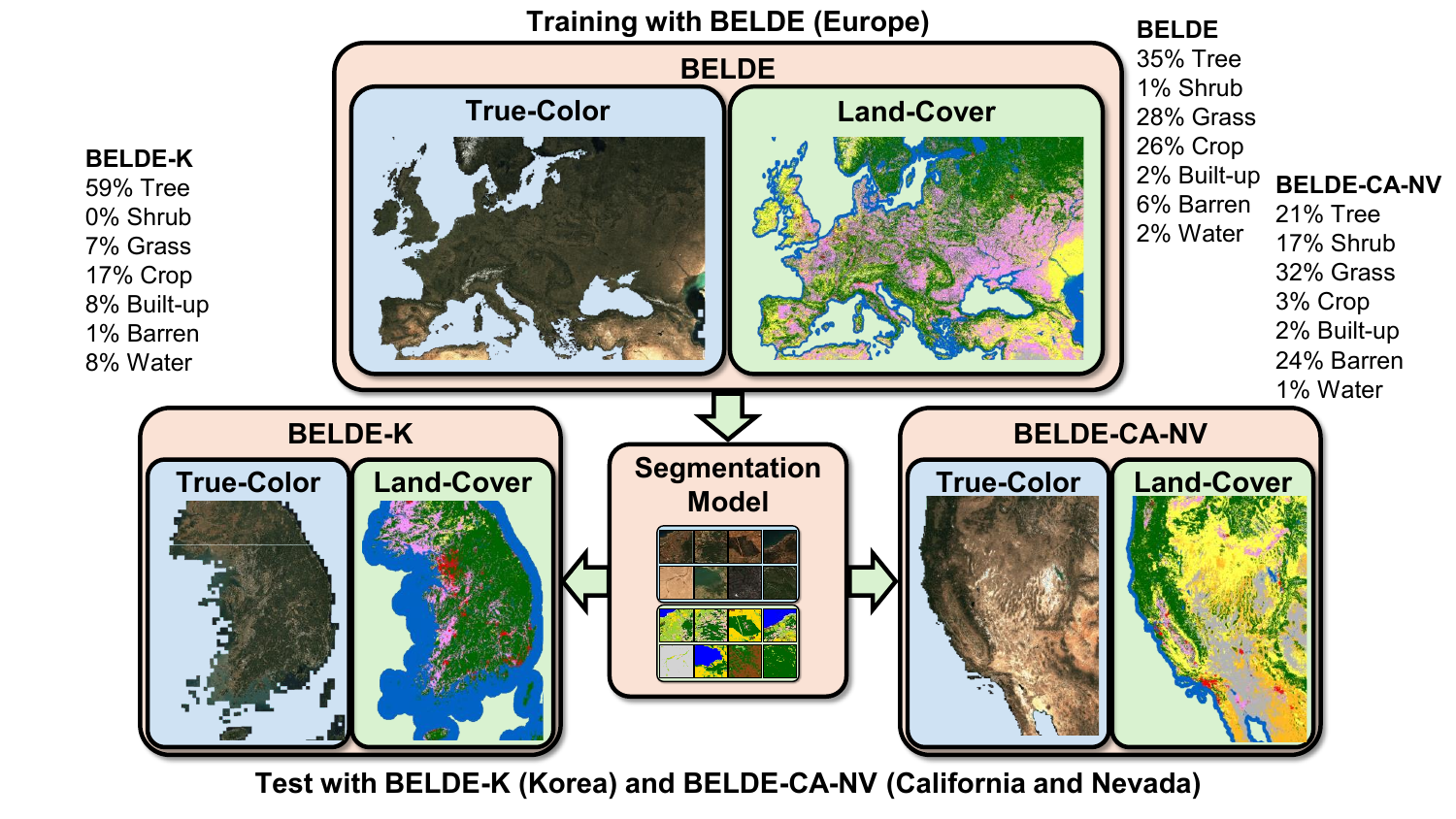}
    \caption{The experimental setup that tests segmentation models trained with BELDE on its extensions, BELDE-K and BELDE-CA-NV.}
    \label{fig:belde_extend}
\end{figure}

\section{Cross-Region Generalization Analysis}

To evaluate the robustness of models trained on BELDE under geographic distribution shift, we introduce two out-of-domain evaluation sets: BELDE-K (Republic of Korea) and BELDE-CA-NV (California and Nevada, USA), illustrated in Figure~\ref{fig:belde_extend}. These datasets are constructed using the same automated pipeline as BELDE, ensuring consistency in resolution, preprocessing, and label space while introducing distinct geographic and ecological distributions.

BELDE-K covers an area of 108,835km$^2$ with 16,607 image-mask pairs, while BELDE-CA-NV spans 577,732km$^2$ with 88,155 pairs. Together, they represent substantially different land-cover distributions compared to the European training domain, enabling systematic evaluation of cross-region generalization in RGB-only Earth observation.

\subsection{Effect of Geographic Domain Shift}

Tables~\ref{tab:beldek} and~\ref{tab:beldecn} summarize model performance when trained on BELDE (Europe) and evaluated on BELDE-K and BELDE-CA-NV, respectively, in zero-shot setting. Across all architectures, we observe a consistent performance drop compared to in-domain evaluation, confirming the presence of significant geographic domain shift in RGB land-cover segmentation, which is also observable from Figure~\ref{fig:composition}. BELDE on average is 35\% tree, 28\% grass, 26\% crop, 6\% barren and only 2\% built-up,  meanwhile BELDE-K is 59\% tree, 17\% crop, 8\% built-up, 8\% water and only 1\% barren, indicating a very different composition of land-cover classes increasing the difficulty of model transfer due to the domain-shift.

On BELDE-K, transformer-based and hybrid architectures achieve the highest performance, with MaxViT and EfficientFormer-L7 reaching F1-scores of 58.2\% and 58.3\%, respectively. CNN-based architectures such as UNet and UNet++ remain competitive, achieving 56.6\% and 56.5\%, while lightweight LALE models provide a more efficient but lower-performing alternative, with LALE-S3 reaching 52.1\%.

A similar trend is observed on BELDE-CA-NV, where MaxViT and EfficientFormer-L7 achieve the highest F1-scores of 66.4\% and 66.2\%. CNN-based models such as UNet and SegFormer achieve mid-range performance around 65\%, while LALE models maintain lower but computationally efficient performance levels.

\subsection{Robustness Across Architectural Families}

While absolute performance varies across architectures, all model families exhibit consistent degradation under domain shift, indicating that geographic variability rather than model capacity is the dominant limiting factor in cross-region generalization.

We further observe that lightweight architectures such as LALE exhibit relatively stable variance across runs, suggesting consistent optimization behavior under distribution shift, despite lower absolute performance. In contrast, larger transformer-based models show higher sensitivity to geographic variation, reflected in increased variance across evaluation runs.

\subsection{Discussion}

These results highlight that performance gains observed in-domain do not directly translate to out-of-domain settings, particularly under strong geographic shifts. This emphasizes the importance of geographically diverse benchmarks such as BELDE for evaluating real-world robustness in Earth observation systems. More broadly, BELDE-K and BELDE-CA-NV suggest that RGB-only segmentation models rely heavily on region-specific spatial and textural cues and that generalization across continents remains a challenging open problem in large-scale land-cover mapping.

\begin{table}[ht]
\centering
\caption{Zero-shot BELDE-K performance (\%) grouped by architectural cohorts.}
\label{tab:beldek}
\renewcommand{\arraystretch}{1.1}
\resizebox{0.8\textwidth}{!}{\begin{tabular}{l|c|c|c|c|c}
Architecture & Accuracy & Precision & Recall & F1-score & IoU \\
\midrule
\rowcolor{gray!12}
DeepLabV3 & $80.4 \pm 0.4$ & $62.3 \pm 0.3$ & $55.1 \pm 0.1$ & $55.3 \pm 0.1$ & $45.6 \pm 0.0$ \\
\rowcolor{gray!12}
DeepLabV3+ & $80.9 \pm 0.5$ & $63.1 \pm 0.3$ & $55.8 \pm 0.5$ & $55.9 \pm 0.7$ & $46.3 \pm 0.6$ \\
\rowcolor{gray!12}
FPN & $80.4 \pm 0.2$ & $64.0 \pm 0.2$ & $55.8 \pm 0.2$ & $55.9 \pm 0.2$ & $46.3 \pm 0.2$ \\
\rowcolor{gray!12}
Linknet & $79.5 \pm 0.5$ & $63.7 \pm 0.2$ & $55.1 \pm 0.4$ & $54.9 \pm 0.6$ & $45.5 \pm 0.5$ \\
\rowcolor{gray!12}
PSPNet & $79.8 \pm 0.4$ & $62.0 \pm 0.1$ & $54.2 \pm 0.1$ & $54.5 \pm 0.1$ & $44.8 \pm 0.2$ \\
\rowcolor{gray!12}
Segformer & $80.2 \pm 0.2$ & $64.4 \pm 0.6$ & $55.3 \pm 0.4$ & $55.4 \pm 0.3$ & $45.9 \pm 0.1$ \\
\rowcolor{gray!12}
Unet & $80.8 \pm 0.1$ & $64.1 \pm 0.2$ & $56.5 \pm 0.5$ & $56.6 \pm 0.5$ & $47.0 \pm 0.4$ \\
\rowcolor{gray!12}
Unet++ & $80.7 \pm 0.6$ & $64.3 \pm 0.5$ & $56.5 \pm 0.4$ & $56.5 \pm 0.7$ & $46.9 \pm 0.7$ \\
\midrule
DeiT3 & $81.4 \pm 0.7$ & $64.9 \pm 0.9$ & $56.8 \pm 0.3$ & $56.9 \pm 0.6$ & $47.3 \pm 0.6$ \\
EfficientFormer-L1 & $81.4 \pm 0.7$ & $64.2 \pm 0.5$ & $57.4 \pm 0.5$ & $57.5 \pm 0.4$ & $47.8 \pm 0.5$ \\
EfficientFormer-L3 & $82.0 \pm 1.0$ & $65.3 \pm 0.5$ & $57.3 \pm 0.8$ & $57.7 \pm 1.0$ & $48.1 \pm 0.9$ \\
EfficientFormer-L7 & $81.9 \pm 0.4$ & $65.0 \pm 0.8$ & $57.9 \pm 0.2$ & $58.3 \pm 0.2$ & $48.5 \pm 0.2$ \\
FastViT-mci0 & $81.9 \pm 0.8$ & $63.9 \pm 1.1$ & $57.3 \pm 0.3$ & $57.2 \pm 1.1$ & $47.5 \pm 1.0$ \\
FastViT-sa12 & $81.7 \pm 0.8$ & $63.9 \pm 1.0$ & $57.6 \pm 0.5$ & $57.3 \pm 0.7$ & $47.8 \pm 0.8$ \\
MaxViT & $82.5 \pm 0.9$ & $64.6 \pm 1.7$ & $57.9 \pm 0.4$ & $58.2 \pm 0.7$ & $48.5 \pm 0.7$ \\
\midrule
\rowcolor{gray!12}
LALE-S3 & $78.6 \pm 0.5$ & $62.7 \pm 0.4$ & $52.5 \pm 0.6$ & $52.1 \pm 0.7$ & $43.1 \pm 0.6$ \\
\rowcolor{gray!12}
LALE-S2 & $77.9 \pm 0.5$ & $61.8 \pm 0.5$ & $52.1 \pm 0.4$ & $51.3 \pm 0.6$ & $42.4 \pm 0.5$ \\
\end{tabular}}
\end{table}
\begin{table}[ht]
\centering
\small
\caption{Zero-shot BELDE-CA-NV performance (\%) grouped by architectural cohorts.}
\label{tab:beldecn}
\renewcommand{\arraystretch}{1.1}
\resizebox{0.8\textwidth}{!}{\begin{tabular}{l|c|c|c|c|c}
Architecture & Accuracy & Precision & Recall & F1-score & IoU \\
\midrule
\rowcolor{gray!12}
DeepLabV3 & $62.8 \pm 2.1$ & $63.8 \pm 2.9$ & $67.5 \pm 1.2$ & $63.4 \pm 2.0$ & $49.9 \pm 2.2$ \\
\rowcolor{gray!12}
DeepLabV3+ & $63.4 \pm 0.3$ & $67.0 \pm 0.7$ & $67.7 \pm 0.4$ & $65.0 \pm 0.3$ & $51.9 \pm 0.3$ \\
\rowcolor{gray!12}
FPN & $63.8 \pm 0.3$ & $65.6 \pm 0.5$ & $67.8 \pm 0.6$ & $64.3 \pm 0.2$ & $51.1 \pm 0.3$ \\
\rowcolor{gray!12}
Linknet & $64.1 \pm 0.3$ & $65.9 \pm 0.7$ & $67.8 \pm 0.2$ & $64.5 \pm 0.3$ & $51.3 \pm 0.4$ \\
\rowcolor{gray!12}
PSPNet & $62.8 \pm 0.6$ & $67.3 \pm 0.6$ & $66.3 \pm 0.7$ & $64.0 \pm 0.5$ & $50.5 \pm 0.6$ \\
\rowcolor{gray!12}
Segformer & $64.0 \pm 0.1$ & $66.7 \pm 0.5$ & $68.2 \pm 0.5$ & $65.0 \pm 0.2$ & $52.1 \pm 0.2$ \\
\rowcolor{gray!12}
Unet & $64.1 \pm 0.5$ & $66.5 \pm 0.4$ & $68.0 \pm 0.4$ & $65.1 \pm 0.5$ & $52.2 \pm 0.4$ \\
\rowcolor{gray!12}
Unet++ & $64.1 \pm 0.9$ & $66.9 \pm 0.5$ & $68.5 \pm 1.0$ & $65.1 \pm 1.0$ & $52.2 \pm 1.1$ \\
\midrule
DeiT3 & $63.2 \pm 0.3$ & $65.4 \pm 2.7$ & $66.6 \pm 1.3$ & $63.6 \pm 1.8$ & $50.2 \pm 2.0$ \\
EfficientFormer-L1 & $63.7 \pm 0.7$ & $66.9 \pm 1.7$ & $67.8 \pm 0.4$ & $64.8 \pm 0.7$ & $51.8 \pm 0.9$ \\
EfficientFormer-L3 & $64.2 \pm 0.4$ & $67.4 \pm 1.6$ & $68.2 \pm 0.2$ & $65.5 \pm 0.6$ & $52.5 \pm 0.6$ \\
EfficientFormer-L7 & $64.9 \pm 1.4$ & $68.7 \pm 0.9$ & $68.8 \pm 1.0$ & $66.2 \pm 0.9$ & $53.6 \pm 1.1$ \\
FastViT-mci0 & $64.8 \pm 0.2$ & $68.0 \pm 0.6$ & $68.6 \pm 0.1$ & $65.7 \pm 0.4$ & $52.8 \pm 0.5$ \\
FastViT-sa12 & $64.7 \pm 0.6$ & $66.5 \pm 1.6$ & $68.6 \pm 0.2$ & $65.3 \pm 1.0$ & $52.4 \pm 1.1$ \\
MaxViT & $65.7 \pm 0.2$ & $67.4 \pm 1.2$ & $69.8 \pm 0.2$ & $66.4 \pm 0.8$ & $53.6 \pm 0.9$ \\
\midrule
\rowcolor{gray!12}
LALE-S3 & $60.7 \pm 0.6$ & $62.6 \pm 1.4$ & $65.6 \pm 0.8$ & $60.9 \pm 1.1$ & $47.1 \pm 1.2$ \\
\rowcolor{gray!12}
LALE-S2 & $59.6 \pm 0.9$ & $62.2 \pm 0.7$ & $64.7 \pm 0.7$ & $60.3 \pm 0.4$ & $46.2 \pm 0.5$ \\
\end{tabular}}
\end{table}

\section{Computational Costs}

\begin{table}[ht]
\centering
\caption{Computational costs of training experiments, optimization and conceptualization which includes architectural search.}
\label{tab:training_summary}
{\footnotesize
\begin{tabular}{lccc}
\hline
\textbf{Experiment Group} & \textbf{Trainings} & \textbf{Hours} & \textbf{\%} \\
\hline
Conceptualization/optimization & 58 & 544  & 51 \\ 
\midrule
Dense Prediction Transformers & 21 & 67 & 6 \\
CNN-based architectures & 24 & 125 & 12 \\
Lightweight architectures & 6 & 13 & 1 \\ 
\midrule
California-Nevada Extension & 51 & 206 & 19 \\ 
Republic of Korea Extension & 51 & 122 & 11 \\ 
\hline
\textbf{Total} & \textbf{211} & \textbf{1,077} & \textbf{100} 
\end{tabular}}
\end{table}

Table~\ref{tab:training_summary} summarizes the computational resources required for all experiments. In total, 211 training runs were conducted on NVIDIA H100 GPUs. An initial set of 58 conceptualization including architectural search and proof-of-concept experiments (544 GPU-hours) were used to optimize the training pipeline, including data loading, CPU worker allocation, and batch-size configuration. The final benchmarking study comprised 21 Dense Prediction Transformer experiments (67 GPU-hours), 24 CNN-based experiments (125 GPU-hours), and 6 lightweight architecture experiments (13 GPU-hours). To evaluate cross-region generalization, an additional 51 training runs were performed for BELDE-K (122 GPU-hours) and 51 training runs for BELDE-CA-NV (206 GPU-hours). All experiments were conducted three times to compute the mean and standard deviation of the results, which form the benchmark values reported throughout the paper. Overall, the experiments required 1,077 GPU-hours, reflecting the computational demands of benchmarking diverse segmentation architectures on a million-sample Earth observation dataset.

\section{Conclusion}
\label{sec:conclusion}

We introduced BELDE, a large-scale, geographically diverse benchmark dataset for RGB land-cover semantic segmentation, consisting of over one million curated image-mask pairs derived from Sentinel-2 and ESA WorldCover data over Europe at 10-meter spatial resolution. To enable systematic evaluation of cross-regional generalization, we further released two out-of-domain extensions: BELDE-K (Republic of Korea) and BELDE-CA-NV (California-Nevada, USA).

We evaluated 17 segmentation architectures spanning CNN-based encoder-decoder models, transformer-based dense prediction networks, and lightweight efficiency-oriented models. Our results establish a strong baseline for RGB-only Earth observation segmentation under large-scale and geographically diverse settings. While transformer-based models generally achieve the highest accuracy, lightweight architectures offer competitive performance with substantially reduced computational cost, making them attractive for resource-constrained deployment scenarios.

Cross-region evaluation demonstrates consistent performance degradation under geographic domain shift, highlighting the challenge of transferring models trained on European land-cover distributions to structurally and ecologically distinct regions. This underscores the importance of geographically diverse benchmarks for evaluating robustness in Earth observation systems.

Future work will extend BELDE to other continents and multi-temporal data to enable land-cover change detection and introduce multi-year observations for temporal segmentation. In addition, incorporating controlled synthetic augmentation strategies may further improve dataset diversity and help reduce domain shift across geographic regions.

\section*{Acknowledgements}
Ümit Mert Çağlar is a beneficiary of the TUBITAK BIDEB 2211-Domestic Graduate Scholarship Program and TUBITAK 2224-A-Grant Program. The experiments reported in this work were fully performed at TUBITAK ULAKBIM, High Performance and Grid Computing Center (TRUBA).

\bibliographystyle{splncs04}
\bibliography{main}
\end{document}